\documentclass[10pt,twocolumn,twoside]{IEEEtran}
\usepackage{amsfonts}
\usepackage{graphicx}
\usepackage{epstopdf}
\usepackage{subcaption}
\usepackage{multirow}
\usepackage{amsmath}
\usepackage{amssymb}
\usepackage{bm}
\usepackage{float}
\usepackage{url}

\newcommand{\eat}[1]{} 
\begin{document}

\title{Parameter Efficient Deep Neural Networks with Bilinear Projections}
 
\author{Litao~Yu, 
	Yongsheng~Gao,~\IEEEmembership{Senior~Member,~IEEE,}
	Jun~Zhou,~\IEEEmembership{Senior~Member,~IEEE,}
	Jian~Zhang,~\IEEEmembership{Senior~Member,~IEEE}
\thanks{Litao Yu (litao.yu@griffith.edu.au), Yongsheng Gao (yongsheng.gao@griffith.edu.au) and Jun Zhou (jun.zhou@griffith.edu.au) are with the Institute for Integrated and Intelligent Systems, Griffith University, Nathan, 4111, QLD, Australia. Jian Zhang (jian.zhang@uts.edu.au) is with Global Big Data Technologies Centre, University of Technology Sydney, Ultimo, 2007, NSW, Australia. This work was supported by the Australian Research Council under Discovery Grants DP140101075 and DP180100958.}
 \thanks{Manuscript received March 4, 2019.}
}
 
\maketitle
 
\markboth{IEEE Transactions on Neural Networks and Learning Systems}%
 {Yu \MakeLowercase{\textit{et al.}}:Parameter Efficient Deep Neural Networks with Bilinear Projections}

\begin{abstract}

Recent research on deep neural networks (DNNs) has primarily focused on improving the model accuracy. Given a proper deep learning framework, it is generally possible to increase the depth or layer width to achieve a higher level of accuracy. However, the huge number of model parameters imposes more computational and memory usage overhead and leads to the parameter redundancy. In this paper, we address the parameter redundancy problem in DNNs by replacing conventional full projections with bilinear projections. For a fully-connected layer with $D$ input nodes and $D$ output nodes, applying bilinear projection can reduce the model space complexity from $\mathcal{O}(D^2)$ to $\mathcal{O}(2D)$, achieving a deep model with a sub-linear layer size. However, structured projection has a lower freedom of degree compared to the full projection, causing the under-fitting problem. So we simply scale up the mapping size by increasing the number of output channels, which can keep and even boosts the model accuracy. This makes it very parameter-efficient and handy to deploy such deep models on mobile systems with memory limitations. Experiments on four benchmark datasets show that applying the proposed bilinear projection to deep neural networks can achieve even higher accuracies than conventional full DNNs, while significantly reduces the model size.\footnote{The demo code is available at \url{https://github.com/yutao1008/Bi-DNNs}.}

\end{abstract}

\begin{IEEEkeywords}
Bilinear Projections; Convolutional Neural Networks; Recurrent Neural Networks.
\end{IEEEkeywords}

\section{Introduction}
\label{sect:introduction}

With the brilliant re-design of neural networks and a significant improvement in the training efficiency with GPUs, deep learning models have gained a huge success in the research areas of computer vision and natural language processing. Training on a collection of large-scale and well-labelled data, deep architectures have the ability to learn features without the need of feature engineering. This property has the most profound effect on the end results of different learning tasks. 

In visual content analysis tasks, convolutional neural networks (CNNs) have achieved the state-of-the-art performance. They can be used in a variety of applications such as object detection \cite{NIPS15:RCNN}, semantic segmentation \cite{CVPR18:DFN_SEGMENTATION}, visual place localisation \cite{CVPR18:DELS3D} and video tracking \cite{CVPR18:LA_TRACKING}. Starting from LeNet for handwritten digits recognition \cite{ICANN:LENET}, CNNs have been evolving very fast in recent years. The deep models pre-trained on very large image dataset (e.g., ImageNet) can be directly used for visual feature processing in most cases. It is commonly believed that the wider and deeper models are, the better accuracies can be achieved, where Inception network \cite{CVPR16:INCEPTION} and Deep Residual network \cite{CVPR16:RESNET} are the most representative architectures. Inception-Residual network is a hybrid architecture to boost the performance for visual pattern recognition \cite{AAAI17:INCEPTION_RES}. Deep models also perform very well on recurrent learning tasks, for example, image captioning \cite{CVPR15:CG}, machine translation \cite{AAAI17:RECONSTRUCTION_TRANSLATION}, named entity recognition \cite{BIO:DL_NER}, Q\&A systems \cite{ACL17:GSN_QA}. In these applications, recurrent networks exhibit temporal dynamic behaviour to analyse the context information. The most used deep recurrent model is the long short-term memory (LSTM) \cite{NC:LSTM}, which has additional stored states implemented by several fully-connected modules, and the memory can be directly controlled by the neural network. 

When large DNNs are designed with a high number of layers and wide structure in each layer, reducing their model sizes to make them more parameter-efficient becomes critical to meet the requirements of many practical applications. So how to design a parameter-efficient learning framework becomes an open problem. Given an equivalent accuracy level, a smaller DNN with fewer parameters has the following advantages: (1) they are less prone to over-fitting; (2) they require fewer hardware communications; (3) they are easy to be transferred among different devices; (4) they can be better deployed for embedded deployment such as FPGA. 

With this in mind, we directly focus on the problem of identifying a proper structure with fewer parameters, but with equivalent accuracy, compared to the well-known models. Specifically, we design a novel microstructure with bilinear projections to replace the conventional full projection modules. Given a dense layer with $D$ input channels and $D$ output channels, applying bilinear projection can reduce the space complexity from $\mathcal{O}(D^2)$ to $\mathcal{O}(2D)$. However, the structured projection has less freedom, leading to the under-fitting issue. This problem can be alleviated by proper scaling up the feature map size, which will be discussed later in this paper. With such settings, the parameter efficiency and model accuracy can be well balanced.

We experimented several well-known deep architectures with the proposed microstructure on four datasets. These datasets covered three image classification tasks and one recurrent image captioning task. The results prove the effectiveness of our method and its generalisation capability to various deep architectures.

The rest of the paper is organised as follows. Section \ref{SEC:RELATED_WORK} introduces the related work. In section \ref{SEC:METHOD} we elaborate the proposed bilinear projection based microstructure that can be used to replace the conventional neural layers. Experimental results and analysis are presented in section \ref{SEC:EXPERIMENT}. Finally, section \ref{SEC:CONCLUSION} concludes the paper.

\section{Related work} 
\label{SEC:RELATED_WORK}

In this section, we review the latest techniques for small-size DNNs. These techniques can be categorised into two disjoint classes: (1) given a large-sized network structure, how to compress the deep model; and (2) how to directly design light-weight deep architectures.

\subsection{Model compression}

Model compression aims to effectively reduce the model size with a tolerable information loss, which is mainly achieved by pruning, low-rank estimation and quantisation. Given a CNN model, filter pruning is the most straight-forward way to reduce the redundancy, since many filters are unnecessary in the online predictions. Along this line, Han et al. proposed to remove the connections with small weights by using sparsity regularisation \cite{NIPS15:LWC}. This strategy is extended by Wen et al. who use structured sparsity learning method \cite{NIPS16:LSS} to regularise filters. Furthermore, the weight pruning technique can be combined with knowledge distillation \cite{ARXIV:AT}. Although filter-level pruning has also been developed \cite{ARXIV:NT, ICCV17:CP},  how to evaluate the importance of filters for specific tasks remains an open problem. Low-rank estimation is another way for deep model compression. The key idea is to apply matrix factorisation to decompose the large weight matrices into small ones \cite{ARXIV:CP,ARXIV:LR}. Quantisation is a general data encoding method that uses the discrete representations to approximate complex structures, which has been used to compress deep learning models \cite{ICCV15:HTC,CVPR16:QCNN,ARIXV:CVQ}. The basic property of model compression is it does not change the deep learning architectures. Given a fixed DNN framework, nearly all of the above approaches lead to accuracy degradation due to the information loss.

\subsection{Light-weight deep architectures}

Simple or light architectures do not necessarily have worse performance than complex deep models. Some researchers pursue the goal of developing better intuitions about the operation and usage of some structural properties to guide the design of light-weight DNNs. Iandola et al. \cite{ARXIV:SQUEEZENET} designed a novel SqueezeNet with 50 times fewer parameters than AlexNet but achieves a comparable accuracy on ImageNet. To reduce the redundancy of parameters, the authors used two separable low-rank convolution filters instead of a full-rank filter, and designed a bottleneck module to reduce the input channels before convolution. Another light-weight deep structure is MobileNet, which is specifically designed for fast computation, but not necessarily for smaller model size \cite{ARIXV:MOBILENET}. Replacing traditional convolution with depth-wise separable convolution, MobileNet is able to halve the trainable parameters and accelerate the computation, although parameter-efficiency is not the priority for the model design. Later, an inverted residual with a linear bottleneck was introduced to improve the performance without increasing the memory cost \cite{CVPR18:MOBILENET}. Recently, Huang et al. proposed the densely connected network (DenseNet) that has very compact layer structure while keeping the model accuracy \cite{CVPR17:DENSENET}. It has a narrow layer-structure and is very parameter-efficient. Further, they adopted filter-pruning in the training procedure to further balance the model size and accuracy \cite{CVPR18:CONDENSENET}. In \cite{ICCV15:CIR}, Cheng et al. proposed to use circulant projections to replace the linear mapping for CNNs to achieve a sub-linear layer size. However, in these models the accuracy can be hardly maintained even if the number of output channels is significantly increased. Furthermore, it only works during the network shrinkage phase, i.e., the output dimension is lower than the input dimension. 

\section{Proposed Method} \label{SEC:METHOD}

\subsection{Bilinear projection for non-linear mapping}

The matrix form is a specific case of multi-way arrays (tensors) data representation. In fact, all kinds of digital images render the 2D structures with multiple channels such as hyperspectral image \cite{JBO:HYPER_IMAGE}. Besides, in processing images, visual feature descriptors are also quantised into a matrix format such as fisher vector (FV) \cite{IJCV:FV} and vector of locally aggregated descriptors (VLAD) \cite{CVPR10:VLAD}. Therefore, matrix data analysis has become one of the fundamental research topics in image processing. With the success of deep neural networks on large-scale image classification, our intuition is to directly apply bilinear projection on computing 2D structure data as an alternative mapping function to make it more parameter-efficient.

The most basic function of neural networks is to apply an affine transformation with an activation function to map a vector from its original data space to a new feature space:
\begin{equation} \label{EQ:FULL}
h(\bm{x}) = \phi(\bm{x}\mathbf{W} + \bm{b}),
\end{equation} 
where $\bm{x}\in\mathbb{R}^D$ is the input feature vector, $\mathbf{W}\in\mathbb{R}^{D\times K}$ is the weight matrix, $\bm{b}\in\mathbb{R}^K$ is the bias vector, and $\phi(\cdot)$ is an element-wise activation function. The computational complexity and space complexity of the above operation is $\mathcal{O}(DK)$. Assume $D<K$, the complexity of such affine transformation is at least $\mathcal{O}(D^2)$.  

We propose to use bilinear projection to replace conventional full linear projection in the basic neural layer structure. When $D$ is the multiplication of two positive integers, i.e., $D=d_1\times d_2$, the vector $\bm{x}$ can be reshaped (either row-wise or column-wise) into a matrix $\mathbf{x}\in\mathbb{R}^{d_1\times d_2}$ with an operation $r$:
\begin{equation} \label{EQ:MAT}
\mathbf{x} = r(\bm{x}).
\end{equation}  
Correspondingly, $\bm{x}$ can be recovered by flattening $\mathbf{x}$:
\begin{equation}
\bm{x}=r^{-1}(\mathbf{x}).
\end{equation}

Similarly, the dimension of the output $K$ can be decomposed by two factors $k_1$ and $k_2$, i.e., $K=k_1\times k_2$. Instead of applying a full weight matrix $\mathbf{W}$ and a bias vector $\bm{b}$ to map the feature vector $\bm{x}$, bilinear projection is to use two small weight matrices $\mathbf{w}_1\in\mathbb{R}^{k_1\times d_1}$, $\mathbf{w}_2\in\mathbb{R}^{d_2\times k_2}$ and a bias matrix $\mathbf{b}=r(\bm{b})\in\mathbb{R}^{k_1\times k_2}$ to map the feature matrix $\mathbf{x}$. So in a single neural layer, the mapping function becomes:
\begin{equation} \label{EQ:BILINEAR}
h(\mathbf{x}) = \phi(\mathbf{w}_1\mathbf{x}\mathbf{w}_2 + \mathbf{b}).
\end{equation} 

The structured bilinear projection is a special case of full linear projection. When the weight matrix $\mathbf{W}$ satisfies the condition $\mathbf{W}=\mathbf{w}_1^{\top}\otimes \mathbf{w}_2$, where $\otimes$ is the Kronecker product, bilinear projection is equivalent to full linear projection:
\begin{align}\label{EQ:BILINEAR_EQ}
h(\bm{x}) &= r^{-1}(\phi(\mathbf{w}_1\mathbf{x}\mathbf{w}_2 + \mathbf{b})) \nonumber \\
&= \phi(r^{-1}(\mathbf{x})(\mathbf{w}_1^{\top} \otimes \mathbf{w}_2) + \bm{b}) \nonumber \\
&= \phi(r^{-1}(\mathbf{x})\mathbf{W} + \bm{b}).
\end{align} 

The space complexity of bilinear projection is $\mathcal{O}(D+K)$ compared to $\mathcal{O}(DK)$ for full linear projection, thus the number of trainable parameters is significantly reduced. However, the ideal conditions $d_1=d_2=\sqrt{D}$ and $k_1=k_2=\sqrt{K}$ can hardly be satisfied, so the bilinear factors should be selected as the closest positive integers. 

Now let's consider the expected output vector $\bm{y}\in\mathbb{R}^K$, which can also be reshaped as $\mathbf{y}=r(\bm{y})\in\mathbb{R}^{k_1\times k_2}$. If we use the mean-squared-error cost as the objective, the stochastic loss function is:
\begin{equation} \label{EQ:LOSS}
 J(\mathbf{w}_1,\mathbf{w}_2,\mathbf{b})=\frac{1}{2}\|h(\mathbf{x})-\mathbf{y}\|^2+ \frac{\lambda}{2}(\|\mathbf{w}_1\|^2+\|\mathbf{w}_2\|^2). 
\end{equation}
 
The first term is a mean-squared-error cost and the second term is a weight decay. Our goal is to minimise $J(\mathbf{w}_1,\mathbf{w}_2,\mathbf{b})$ as a function of $\mathbf{w}_1,\mathbf{w}_2$ and $\mathbf{b}$. To train the single layer, these parameters are initialised with small random values with a pre-defined data distribution, e.g., zero-centred normal distribution with 0.01 covariance, then a batch gradient descent algorithm is applied in the back-propagation. One iteration of gradient descent updates $\mathbf{w}_1,\mathbf{w}_2$ and $\mathbf{b}$ as follows:
\begin{align} \label{EQ:GRADIENT}
\mathbf{w}_1 &\rightarrow \mathbf{w}_1 - \eta \frac{\partial J}{\partial \mathbf{w}_1}, \\
\mathbf{w}_2 &\rightarrow \mathbf{w}_2 - \eta \frac{\partial J}{\partial \mathbf{w}_2}, \\
\mathbf{b} &\rightarrow \mathbf{b} - \eta \frac{\partial J}{\partial \mathbf{b}},
\end{align}
where $\eta$ is the learning rate. The partial derivatives of the objective function in Eq. (\ref{EQ:LOSS}) with respect to $\mathbf{w}_1,\mathbf{w}_2$ and $\mathbf{b}$ are computed as follows:
\begin{align}
\frac{\partial J}{\partial \mathbf{w}_1}&=((h(\mathbf{x})-\mathbf{y})\circ \nabla_{\phi})(\mathbf{x}\mathbf{w}_2)^{\top} + \lambda \mathbf{w}_1, \\
\frac{\partial J}{\partial \mathbf{w}_2}&=(\mathbf{w}_1\mathbf{x})^{\top}((h(\mathbf{x})-\mathbf{y})\circ \nabla_{\phi}) + \lambda \mathbf{w}_2, \\
\frac{\partial J}{\partial \mathbf{b}}&=(h(\mathbf{x})-\mathbf{y})\circ \nabla_{\phi}.
\end{align}
where $\nabla_{\phi}$ is the derivative of the activation function, and $\circ$ is the element-wise multiplication. 

To train a single layer with bilinear projection, we can now repeatedly take steps of gradient descent to reduce the value of the cost function in Eq. (\ref{EQ:LOSS}) by feeding data batches into the layer. In a multi-layer neural network, the gradients of different layers are computed with the chain rule.  

\subsection{Comparing bilinear projection with other structured mappings}

In signal processing, structured mapping functions are usually used to encode or decode signals to reduce the parameter redundancy. To mimic the unstructured full projection, there are various structured mapping functions. For example, we can use Hadamard matrices along with a sparse Gaussian matrix for the fast Johnson-Lindenstrauss transform \cite{SOC06:FJLT, KDD11:FLSH}. Such mapping can be used for the embedding is at least Lipschitz, which is at most equivalent to an orthogonal projection. Similarly, circulant projection can replace full projections, and it has been used for binary embedding \cite{ICML14:CBE} and video encoding \cite{CVPR13:CTE}. In the neural network design, circulant projection can sometimes replace the conventional full projections to reduce the parameter redundancy \cite{ICCV15:CIR}. To achieve this, Discrete Fourier Transformation (DFT) and its inverse (IDFT) are applied to facilitate the computation of gradient descent. Thus, the layers can be optimised in the frequency domain rather than in the time domain. 

However, both fast Johnson-Lindenstrauss transform and circulant projection working on large-scale training data tend to become bogged down very quickly as dimension increases \cite{STOC97:TA}, so they are only available in the network shrinkage phase or orthogonal-like mappings, i.e., the dimension of the output is no higher than the input of a layer. For a fully-connected layer without bias, assume the input dimension is $D$ and the output dimension is $K$, applying unstructured mapping requires the matrix $\mathbf{W}\in \mathbb{R}^{D\times K}$, which has the freedom degree of $DK$. Using bilinear projection by replacing $\mathbf{W}$ with two smaller matrices $\mathbf{w}_1$ and $\mathbf{w}_2$, the freedom degree becomes $2\sqrt{DK}$. In fast Johnson-Lindenstrauss transform and circulant projection, the structured mapping matrix is derived from a vector $\mathbf{r}\in\mathbb{R}^{D}$, which has the freedom degree of $D$. Based on the above analysis, only when $D\ge 4K$, the above two mappings can have the same freedom degree with bilinear projection. Consequently, when the forward propagation is in the expansion phase, i.e., the lower-dimensional input is mapped to a higher-dimensional feature space for non-linear mapping, neither fast Johnson-Lindenstrauss transform nor circulant mapping can be used to replace the full projection. In contrast, bilinear projections are much more flexible, which can be used in both shrinkage and expansion phase in the forward pass. 

We now illustrate how to equip popular neural layers with bilinear projection in DNNs for different learning tasks.

\subsection{Fully-connected (dense) layers with bilinear projection} \label{SEC:FC}

Multi-layer neural networks usually contain at least two fully-connected layers: the first layer maps the input variable to a latent feature space, and the last layer is a classifier or a regressor {\em w.r.t.} the objective values. It is commonly known that more hidden nodes can better fit the non-linearity in the latent feature space, yet more computational resource is required. Intuitively, we can directly use bilinear projection with the minimum number of parameters to replace full linear projection. However, the structured mapping in an intermediate layer has less freedom compared to the unstructured mapping, leading to the under-fitting issue. Thus, we introduce a parameter $\alpha$ to scale up the number of output nodes, i.e., the output dimension of the layer is set to $\alpha K$ for bilinear projection. If the number of output channels is multiplied by $\alpha$, the layer size with bilinear projection becomes $\mathcal{O}(D+\alpha K)$, which is still sub-linear to the space complexity of full projection. Scaling up the number of output dimensionality in bilinear projection is equivalent to the multiple rank strategy used for 2D classification \cite{PR:MRSTM} or regression \cite{TIP:MRR}. Given a proper $\alpha$, the number of trainable parameters can be substantially decreased while keeping sufficient freedom to fit the potential data distribution.

Fully-connected layers have huge numbers of parameters in some early CNN architectures. For example, the three fully-connected layers occupy 89\% of the total trainable parameters in the whole VGG19 network \cite{ARXIV:VGG}, which means it is very parameter-redundant. Considering an intermediate layer with 4,096 nodes for both input and output, it requires $4,096\times(4,096+1)=16,784,312$  parameters when applying the full linear projection on it. If we use bilinear projections and set $\alpha=1$,  the number of trainable parameters is substantially reduced to $2\times64\times64+4,096=12,288$. Even if we scale up the output dimension of the intermediate layer, e.g., setting $\alpha=3$, the total number of trainable parameters is only $64\times64+64\times64\times3+4,096\times3=28,672$, achieving an over 99.8\% reduction of the layer size. In recent CNN models such as residual networks \cite{CVPR16:RESNET, ECCV16:RESNET} and dense nets \cite{CVPR17:DENSENET}, fully-connected layers are replaced by global average pooling layers.

\subsection{Convolution layers with bilinear projection}

Natural objects such as images and videos have the property of ``stationary'', which means the statistics of one patch from one object are the same as other patches. The basic idea of convolution operation is to apply $K$ small filters on all possible patches to obtain a feature map, so convolution is a parameter sharing scheme to control the model size, which also alleviates the over-fitting issue effectively. A convolution filter is essentially a small affine transformation with an activation function that processes a small fixed patch as the receptive field. 

Here we only focus on the application of bilinear projection on the 2D convolution layer, which is one of the most basic operators in image processing. The 1D and 3D convolution layers can follow the same rules for the parameter-efficient layer design. In a 2D convolution layer, assume a receptive field in an input image is $w\times h\times c$, where $w$, $h$ and $c$ are the width, the height and the number of channels of the receptive field, respectively. Thus we can consider each receptive field is essentially a small data cubic with the shape $w\times h\times c$. Applying a convolution kernel with $c'$ filters on an image patch means the number of channels is extended from $c$ to $c'$ to acquire new knowledge, where each channel in $c'$ acts as a particular interpretation of the receptive field. In this mapping, convolution operation is to use a weight matrix $\mathbf{W}\in\mathbb{R}^{(w\times h\times c)\times(w\times h\times c')}$ to map the flattened 2D receptive field to a new space then reshape the output vector to a new data cubic, so basically the convolution mapping is a {\em flatten $\rightarrow$ affine transformation $\rightarrow$ reshaping} process. 

A typical CNN consists of convolution layers, down-sampling (pooling) layers and optional fully-connected layers. The shallower convolution layers focus on detecting certain edges, textures and fundamental shapes, while the deeper convolution layers mainly detect more abstract and more general features. Usually, more filters are set in the deeper convolution layers to capture the abstract semantic properties to facilitate the classification. 

We use the bilinear projection in convolution layers in the same way as that in fully-connected layers, except a slightly different setting of the scaling parameter $\alpha$. To improve the performance, we apply $\alpha$ on both small weight matrices in the bilinear projection, i.e., in a single convolution layer, $\mathbf{w}_1\in\mathbb{R}^{\alpha k_1\times d_1}$ and $\mathbf{w}_2\in\mathbb{R}^{d_2\times\alpha k_2}$, respectively. When $\alpha>1$, the convolution layer has more freedom to fit the potential data distribution, but it also increases the computational cost (FLOPs and memory usage) because the intermediate feature size is larger.

Note that in \cite{CVPR15:BICNN}, the authors named their model as ``bilinear CNN'' for fine-grained visual recognition, which is a macro deep architecture that comprises of two parallel CNN streams. Their model is completely different from our microstructure for the layer-design of neural networks.  

\subsection{Word embedding layers with bilinear projection}

Word embedding is one of the most popular representation of document vocabulary. It is capable of capturing context of a word in a document, semantic and syntactic similarity, relation with other words, etc. In natural language processing, learning word embeddings is implemented by Word2Vec model \cite{NIPS13:WORD2VEC}. With the consideration of the document or sentence context, the objective of Word2Vec is to have words with similar context occupy close spatial positions. The word embedding layer is often used as the first layer in a DNN to process the discrete word sequences (sentences), which turns the word indices into dense vectors with a fixed size. The basic operation behind the embedding layer is essentially a linear mapping function, which is to project the sparse and high dimensional vectors into a dense and low dimensional feature space. Thus, bilinear projection can be easily used to alternate the full linear mapping in this case, i.e., the large kernel matrix for linear mapping can be simply calculated by the Kronecker product of two small matrices according to Eq. (\ref{EQ:BILINEAR_EQ}).  

\subsection{Recurrent layers with bilinear projection} 

In some machine learning tasks, time-distributed patterns are usually learned via Recurrent Neural Network (RNN), which is a natural extension of feed-forward networks on modeling sequence. Here we elaborate on how to use bilinear projections in Long Short-Term Memory (LSTM), which is known to learn patterns with wider ranges of temporal dependencies. 

The core part of LSTM is a memory cell $\bm{c}_t$ at the time step $t$ that records the history of input sequence observed up to that time step \cite{ARXIV:LSTM}. The behaviour of the cell is controlled by three gates computed by fully-connected layers: an input gate $\bm{i_t}$, a forget gate $\bm{f}_t$ and an output gate $\bm{o}_t$. These layers control whether to forget the current cell value if it should read its input and whether to output a new cell value. Specifically, the input gate $\bm{i}_t$ controls whether LSTM considers the current input $\bm{x}_t$, the forget gate $\bm{f}_t$ controls whether LSTM forgets the previous memory $\bm{c}_{t-1}$, and the output gate $\bm{o}_t$ controls how much information will be read from memory $\bm{c}_t$ to the current hidden state $\bm{h}_t$. The definition of these gates, the cell update and output are as follows:
\begin{align}
\bm{i}_t & = \sigma(\bm{x}_t\mathbf{W}_{ix}+\bm{h}_{t-1}\mathbf{W}_{ih}+\bm{b}_i),  \\
\bm{f}_t & = \sigma(\bm{x}_t\mathbf{W}_{fx}+\bm{h}_{t-1}\mathbf{W}_{fh}+\bm{b}_f), \\
\bm{o}_t & = \sigma(\bm{x}_t\mathbf{W}_{ox}+\bm{h}_{t-1}\mathbf{W}_{oh}+\bm{b}_o), \\
\bm{g}_t & = \tanh (\bm{x}_t\mathbf{W}_{gx}+\bm{h}_{t-1}\mathbf{W}_{gh}+\bm{b}_g),  \\
\bm{c}_t & = \bm{f}_t \circ \bm{c}_{t-1} + \bm{i}_t\circ \bm{g}_t,  \\  
\bm{h}_t &= \bm{o}_t \circ \tanh (\bm{c}_t),
\end{align}
where $\sigma(\cdot)$ is the sigmoid function. The weight matrices $\mathbf{W}_{*x}$ and $\mathbf{W}_{*h}$ are the LSTM state and recurrent transformations, and $\bm{b}_*$ are bias vectors. An LSTM needs 8 full projection matrices, which is not paratemer-efficient. To implement an LSTM with bilinear projections, we first transform the input vector $\bm{x}_t$ at the time step $t$, and use the matrix representations of the hidden state and all gates, then replace $\mathbf{W}_*$ and $\bm{b}_*$ with $\mathbf{w}_*^{(1)}$, $\mathbf{w}_*^{(2)}$ and $\mathbf{b}_*$, respectively. The new computation stream becomes:  
\begin{align}
\mathbf{i}_t & = \sigma(\mathbf{w}^{(1)}_{ix}\mathbf{x}_t \mathbf{w}^{(2)}_{ix}+\mathbf{w}^{(1)}_{ih}\mathbf{h}_{t-1}\mathbf{w}^{(2)}_{ih}+\mathbf{b}_i), \\
\mathbf{f}_t & = \sigma(\mathbf{w}^{(1)}_{fx}\mathbf{x}_t \mathbf{w}^{(2)}_{fx}+\mathbf{w}^{(1)}_{fh}\mathbf{h}_{t-1}\mathbf{w}^{(2)}_{fh}+\mathbf{b}_f), \\
\mathbf{o}_t & = \sigma(\mathbf{w}^{(1)}_{ox}\mathbf{x}_t \mathbf{w}^{(2)}_{ox}+\mathbf{w}^{(1)}_{oh}\mathbf{h}_{t-1}\mathbf{w}^{(2)}_{oh}+\mathbf{b}_o), \\
\mathbf{g}_t & = \tanh (\mathbf{w}^{(1)}_{gx}\mathbf{x}_t \mathbf{w}^{(2)}_{gx}+\mathbf{w}^{(1)}_{gh}\mathbf{h}_{t-1}\mathbf{w}^{(2)}_{gh}+\mathbf{b}_g),  \\
\mathbf{c}_t & = \mathbf{f}_t \circ \mathbf{c}_{t-1} + \mathbf{i}_t\circ \mathbf{g}_t,  \\  
\mathbf{h}_t &= \mathbf{o}_t \circ \tanh (\mathbf{c}_t). 
\end{align}

Similar to the fully-connected layer introduced in Section \ref{SEC:FC}, we use the scaling parameter $\alpha$ to control the size of the hidden state vector $\bm{h}_t$, which can be recovered by $\mathbf{h}_t$.

\section{Experiments and analysis} \label{SEC:EXPERIMENT}

We apply bilinear projection on different neural layers to build deep learning models, then test their performances for image classification and image captioning tasks on four public datasets. Although it is not the main goal to obtain the state-of-the-art accuracies on these datasets, we empirically show that with a proper scaling up of the layer size, DNNs with bilinear projections can still outperform those using traditional unstructured full projections.

\subsection{Datasets}

We conducted experiments of image classification on ILSVRC ImageNet 2012, CIFAR-10 and SVHN datasets. 

The ILSRVC ImageNet is one of the largest image datasets for classification and object localisation, which contains 1,281,167 and 50,000 samples for training and validation, respectively. All images in this dataset are with high resolution and are well labelled with 1,000 categories. 

The CIFAR-10 dataset is a subset of 80M tiny image collections, which consists of 60,000 low-resolution colour images with $32\times32$ pixels. The training and testing sets contain 50,000 and 10,000 images respectively. In the training procedure, we held out 5,000 images in the training set as a validation set. 

The Street View House Numbers (SVHN) dataset contains $32\times32$-pixel coloured digit images. In the whole set, 73,257 images are used for training and 26,302 images are used for testing, respectively. In our experiment, we randomly selected 6,000 images from the training set for validation.

To test the effectiveness of LSTM with bilinear projections, we conducted the experiment of recurrent image captioning on Flickr8k dataset \cite{HLT10:FLICKR8K}. This dataset is a benchmark for sentence-based image description, which contains 6,000 images for training, 1,000 images for validation, and 1,000 images for testing. Each image has been annotated with 5 sentences that are relatively visual and unbiased. 

\subsection{Models} 

On the ILSRVC ImageNet dataset, we tested the VGG19 model \cite{ARXIV:VGG}. Since it is extremely time-consuming to train the VGG19 model from random initialisation of parameters, we used the pre-trained convolution models but only optimised the 2 fully-connected layers.  

On CIFAR-10 and SVHN datasets, we tested four different deep classification models. Specifically, we tested two ``heavy'' networks, a small VGG net (S-VGG) and deep residual network (ResNet-56), and two ``light'' models, SqueezeNet \cite{ARIXV:MOBILENET} and MobileNet v2 \cite{CVPR18:MOBILENET}, respectively. 

We built a small VGG net consisting of 3 convolution blocks, 1 fully-connected layer and 1 soft-max layer. Each convolution block has 3 convolution layers and a max-pooling layer for down-sampling. Following the 3 convolution blocks, the feature maps are flattened then connected with a 1,024-D dense layer. The last layer is with the soft-max activation, and the output dimension is the same with the number of classes. 

The idea of ResNets is to learn the addictive residual functions {\em w.r.t} the identity mappings, which is implemented by attaching identity skip connections. We followed \cite{ECCV16:RESNET} to construct the ResNet-56 network. It begins with a residual block, followed by 3 computational stages, each of which containing 9 residual blocks. At the beginning of the stages, the feature map size is down-sampled by a convolution layer with $\text{strides}=2$, while the number of filter maps is up-sampled. Within each stage, the layers have the same number of filters and the same filter map sizes. Following the residual blocks, a batch-normalisation layer with the Rectified Linear Unit (ReLU) activation, and a 2D average pooling layer are applied. The last soft-max layer is for classification.  

We also experimented with two light-weight deep models: SqueezeNet \cite{ARIXV:MOBILENET} and MobileNet v2 \cite{CVPR18:MOBILENET}. Both models are designed for efficient computation. We constructed a small SqueezeNet with 5 fire modules. Each module contains 3 convolution layers with ReLU activations. The MobileNet v2 was built in the same way as it was introduced in the original paper, and both of its width and depth multipliers were set to 1. Applying bilinear projections on these two light deep networks can further reduce the models and boost the computational efficiency.

The Recurrent Image Captioning model was proposed in \cite{CVPR15:CG}, which is comprised by three components: a CNN image encoder, a natural language processor and a recurrent decoder. We used the pre-trained VGG19 model (the last soft-max layer is removed) as the image encoder, with the output dimension 4,096. The natural language processor contains a word embedding layer, an LSTM layer, and a time-distributed dense layer. In the decoder, an LSTM is employed to generate captions. It takes the image feature vector and partial captions at the current time-step as input,  and generates the next most probable word as output.

\subsection{Implementation details, evaluation metrics and experimental settings}

We implemented the bilinear projection based fully-connected layer, word embedding layer, 2D convolution layer and LSTM layer with Keras backend on Tensorflow, by replacing the full weight matrix with two small matrices. Specifically, the bilinear projection was implemented by two tensor dot operations. The APIs of the layers with bilinear projections can be used in a similar way as the original ones, so all macrostructures of these models were kept without any changes. An additional scale hyper-parameter $\alpha$ can be optionally set to control the output feature map size. We used the models with full projections as benchmarks, and set different values of $\alpha$ ranging from 1 to 3 for bilinear projections, until the best performance was met. We found that simply setting $\alpha=3$, the deep models with structured bilinear projections can achieve the same or even higher accuracies than that of unstructured full projections. We also conducted the experiment using channel pruning (CP) \cite{ICCV17:CP} and low-rank expansion (LR) \cite{ARXIV:LR} to reduce the parameter redundancy in convolution layers thus accelerate the computation. 

We used accuracy to evaluate the performance of image classification on ILSVRC ImageNet, CIFAR-10, and SVHN datasets. For image captioning, we used BLEU score \cite{ACL02:BLEU}, which is the most commonly used metric in image description tasks.

For image preprocessing, we normalised the data using the channel means and standard deviations. In the training process, we adopted online image augmentation methods including random rotation, cropping, flipping, channel swapping, etc. In image classification tasks, all models were trained using Adam optimiser with the learning rate 0.001, $\beta_1=0.9$, and $\beta_2=0.999$. In the image captioning task, we applied the RMSprop optimiser with the learning rate of 0.001 and $\rho=0.9$. We set different batch sizes ranging from 32 to 256 for the models to maximise the usage of GPU memory. We used the categorical cross-entropy as the loss function, and chose the best models with the lowest cross-entropy values on validation sets for model selection. 

Our experiments were conducted on a workstation equipped with two NVIDIA Titan Xp GPU cards.

\subsection{Model size comparisons}

\begin{table*}[t]
\centering \small
\caption{Summary of trainable parameters in different deep models. The EXC columns exclude the last soft-max layer, and the INC columns show the total numbers of trainable parameters in the models. }
\label{TB:NTP}
\begin{tabular}{|c|rr|rr|rr|rr|}
\hline
\multirow{3}{*}{Models} &  \multicolumn{2}{c|}{\multirow{2}{*}{ Full}}   & \multicolumn{6}{c|}{Bilinear projection (BP)} \\
\cline{4-9}  & &   & \multicolumn{2}{c|}{$\alpha=1$}     & \multicolumn{2}{c|}{$\alpha=2$}   & \multicolumn{2}{c|}{$\alpha=3$}    \\
\cline{2-9}  & \multicolumn{1}{c}{EXC}     & \multicolumn{1}{c|}{INC}   & \multicolumn{1}{c}{EXC}     & \multicolumn{1}{c|}{INC}   & \multicolumn{1}{c}{EXC}     & \multicolumn{1}{c|}{INC}   & \multicolumn{1}{c}{EXC}     & \multicolumn{1}{c|}{INC}  \\
\hline
S-VGG		       & 2,578,944  & 2,589,194  & 8,548  & 18,798  & 29,928  & 50,418  & 57,996  & 88,726       \\
\hline
ResNet-56		& 1,660,768  & 1,663,338  & 32,128  & 34,698  & 128,416  & 138,666  & 299,456  & 309,706 \\
\hline
SqueezeNet		& 467,370  & 469,380  & 10,080  & 15,210  & 40,128  & 60,618  & 89,196  & 135,286  \\
\hline
MobileNet v2		& 2,223,872  & 2,236,682  & 112,432  & 125,242  & 654,056  & 705,266  & 3,065,346  & 3,180,556   \\
\hline
Image captioner	& 7,712,160  & 15,976,416  & 26,468  & 8,290,724  & 50,224  & 16,570,480  & 70,848  & 24,847,104   \\
\hline
\end{tabular}
\end{table*}

The static model size of DNN is mainly dependent on the number of trainable parameters. In Table \ref{TB:NTP}, we summarise the numbers of trainable parameters of the small DNNs we tested. From the table, we see that applying bilinear projections (BP) on CNNs can significantly reduce the number of parameters. For example, when $\alpha=3$, applying bilinear projections, the S-VGG net and ResNet-56 are 29.2x and 5.4x more parameter-efficient, respectively. Also, heavier CNNs with bilinear projections are substantially smaller than SqueezeNet and MobileNet v2. From this perspective, using sub-linear layer structures needs fewer trainable parameters than designing a macro light-weight deep architecture. When $\alpha=3$, two models MobileNet v2 and image captioner implemented by bilinear projections, are less parameter-efficient than original models with full projections. In MobileNet v2, the depth-wise convolution (as a generic and low-level operation in Tensorfow) was not implemented in our settings. Consequently, when scaling up the feature map size in convolution layers, the kernel mapping sizes in depth-wise convolution layers are also significantly increased. In the image captioning model, the vocabulary size is 8,256, leading to a very large soft-max layer for word generation. Ignoring the last soft-max layer, applying bilinear projections for word embedding, LSTM and dense layers is 153.5x and 108.8x more parameter-efficient when $\alpha=2$ and $\alpha=3$, respectively. However, considering the doubled output dimension of LSTM, the whole size of the model is slightly larger. 

\subsection{Experimental results on image classification}

\subsubsection{Results on ImageNet dataset}
\begin{figure*}[t]
\centering
    \begin{minipage}{0.4\textwidth}
        \includegraphics[width=1\textwidth]{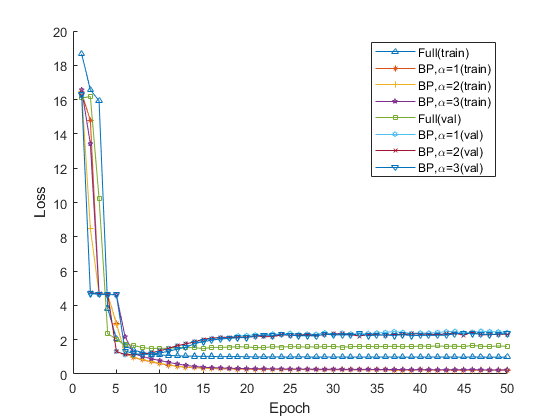}
	\subcaption{Loss}
    \end{minipage}
    ~ 
    \begin{minipage}{0.4\textwidth}
        \includegraphics[width=1\textwidth]{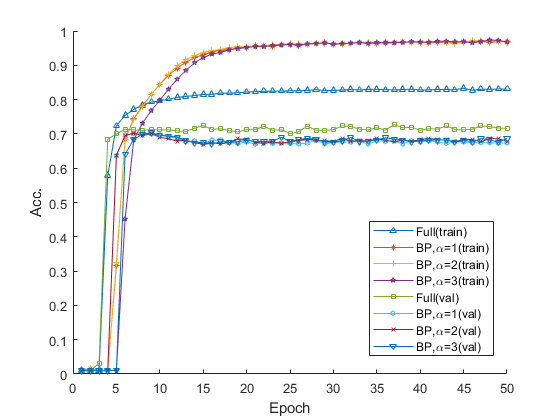}
        \subcaption{Accuracy}
    \end{minipage}
    \caption{Loss and accuracy curves on ImageNet dataset. }
\label{FIG:ILA}
\end{figure*}

We show the loss and accuracy curves of various VGG19 models in Fig. \ref{FIG:ILA} and display the validation errors in Table \ref{TB:ERR_IMAGENET}. We can see that the optimisation significantly improves the accuracies of both unstructured full projections and structured bilinear projections. The performance of VGG19 with bilinear projections on the two dense layers is quite competitive compared against the original model, yet with fraction of the space cost. By tweaking the structure to increase the value of $\alpha$, the proposed bilinear dense layers take only a marginally larger space, but lower the top-1 error rate to 29\% and the top-5 error rate to 10\%, respectively. 

\begin{table}[t]
\centering \small
\caption{Validation error rates of VGG19 on ImageNet.}
\label{TB:ERR_IMAGENET}
\begin{tabular}{|c|c|c|c|c|}
\hline
\multirow{2}{*}{Models}   & \multirow{2}{*}{Full}     & \multicolumn{3}{c|}{BP}    \\
\cline{3-5} & & $\alpha=1$ & $\alpha=2$ & $\alpha=3$\\
\hline
Top-1 error     & 28.5  & 36.7  & 32.2 & 28.9     \\
Top-5 error	& 9.4  & 14.3   & 11.9 & 10.2        \\
\hline
\end{tabular}
\end{table}

\begin{figure*}[t]
\centering
    \begin{minipage}{0.4\textwidth}
        \includegraphics[width=1\textwidth]{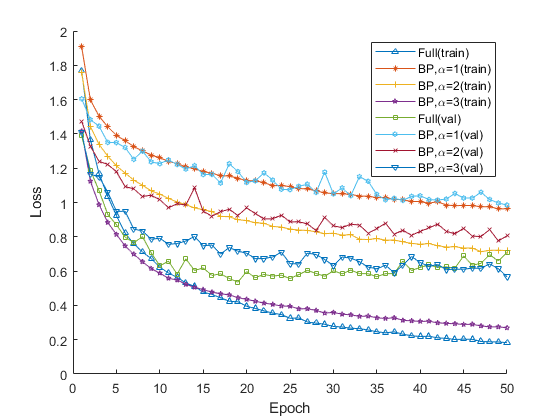}
        \subcaption{Loss}
    \end{minipage}
    ~ 
    \begin{minipage}{0.4\textwidth}
        \includegraphics[width=1\textwidth]{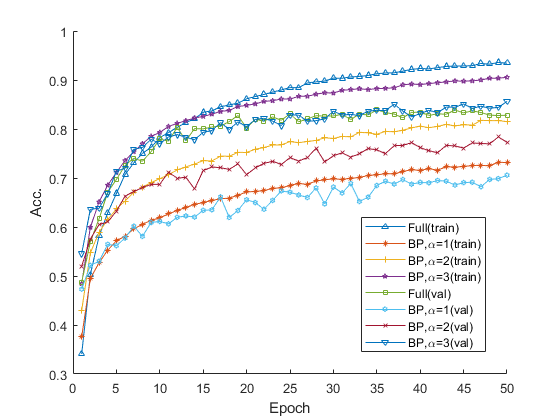}
        \subcaption{Accuracy}
    \end{minipage}
    \caption{Loss and accuracy curves of S-VGG on CIFAR-10 dataset.}
\label{FIG:CL}

\begin{minipage}{0.4\textwidth}
        \includegraphics[width=1\textwidth]{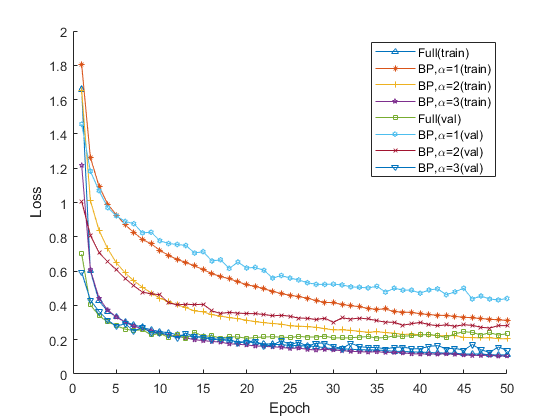}
        \subcaption{Loss}
    \end{minipage}
    ~ 
    \begin{minipage}{0.4\textwidth}
        \includegraphics[width=1\textwidth]{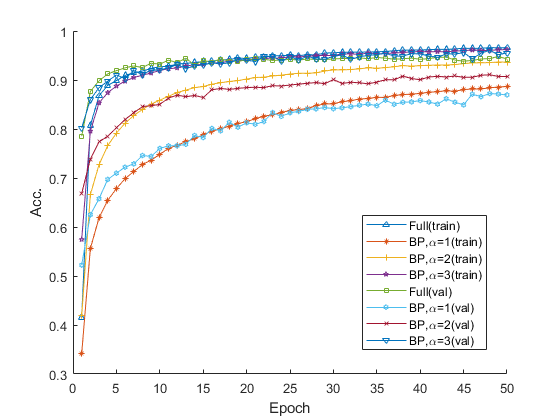}
        \subcaption{Accuracy}
    \end{minipage}
    \caption{Loss and accuracy curves of S-VGG on SVHN dataset.}
\label{FIG:CA}
\end{figure*}

\begin{figure*}[t]
\centering
\begin{minipage}{0.4\textwidth}
        \includegraphics[width=1\textwidth]{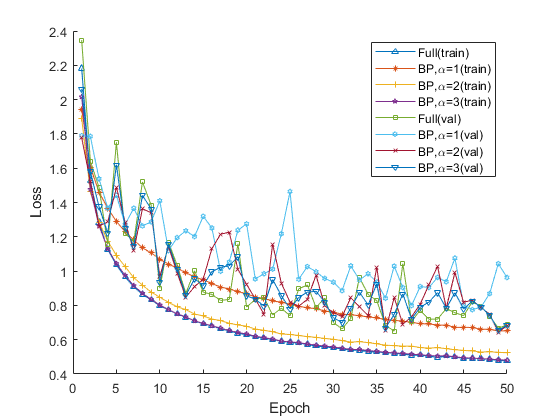}
        \subcaption{Loss}
    \end{minipage}
    ~ 
    \begin{minipage}{0.4\textwidth}
        \includegraphics[width=1\textwidth]{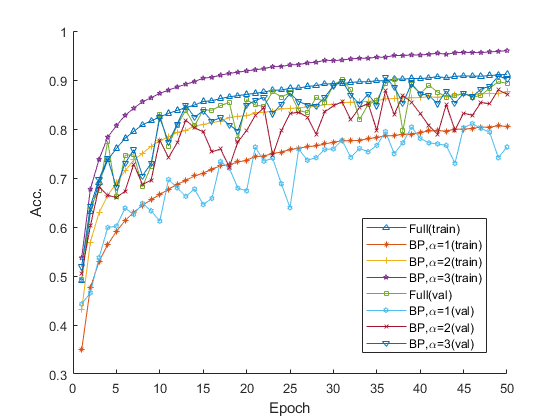}
        \subcaption{Accuracy}
    \end{minipage}
    \caption{Loss and accuracy curves of ResNet-56 on CIFAR-10 dataset.}
\label{FIG:SL}

\begin{minipage}{0.4\textwidth}
        \includegraphics[width=1\textwidth]{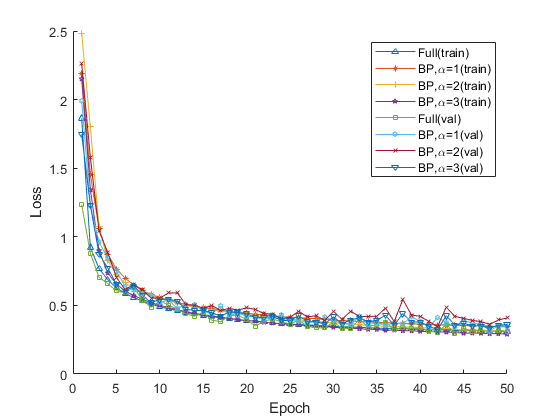}
        \subcaption{Loss}
    \end{minipage}
    ~ 
    \begin{minipage}{0.4\textwidth}
        \includegraphics[width=1\textwidth]{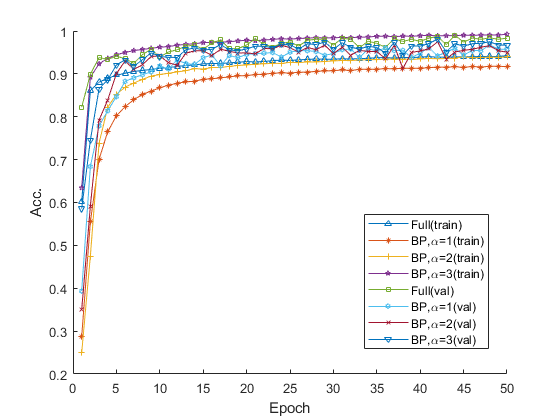}
        \subcaption{Accuracy}
    \end{minipage}
    \caption{Loss and accuracy curves of ResNet-56 on SHVN dataset.}
\label{FIG:SA}
\end{figure*}

\subsubsection{Results on CIFAR-10 and SVHN datasets}
From Fig. \ref{FIG:CL} to Fig. \ref{FIG:SA} we show the training and validation curves within the first 50 training epochs of S-VGG net and ResNet-56 on CIFAR-10 and SVHN datasets, respectively. The loss and accuracy curves of the models with full projections achieve the fastest convergence on the two training sets, while the models with bilinear projections need more training epochs. ResNet-56 is less prone to overfitting compared to S-VGG. In comparison, the gap between training loss and validation loss is smaller with bilinear projections. This is partly because the less freedom of structured mapping also inhibits the over-fitting issue. If we set $\alpha=1$ for ``extreme compression'', all models with bilinear projections generally under-fit on both datasets, with inferior validation accuracies than those of full models. This signifies that we need to scale up the intermediate dimensions for more freedom of the feature representations. When $\alpha=3$ for S-VGG and ResNet-56, the validation loss of the models with bilinear projections is quite close or even smaller than the ones of original models, so it is expected our reconstructed models with proper intermediate outputs controlled by $\alpha$ can have a similar performance on test sets. 

We thus continued increasing the value of $\alpha$ to 4, 5 and 6 to test the performance of S-VGG. The accuracies are 83.1\%, 83.2\% and 83.2\% on CIFAR-10 dataset, and 93.6\%, 93.7\% and 93.7\% on SVHN dataset, respectively. These results signify further scaling up the feature map size to increase the number of channels only marginally improve the model performance. We also observed that when $\alpha\ge 3$, optimising very deep CNNs with bilinear projections becomes very difficult. This bottleneck is further discussed in Section \ref{SUBSEC:DISCUSS}.   

In Table \ref{TB:ACC_CIFAR} and Table \ref{TB:ACC_SVHN}, we show the test accuracies of four deep models, S-VGG, ResNet-56, SqueezeNet and MobileNet v2, on CIFAR-10 and SVHN datasets, respectively. The expected results of bilinear projections with more freedom have been demonstrated for the four deep models we tested. For the rest three models with bilinear projections, the accuracies increase when we set higher intermediate output dimensions on the two datasets. Empirically, when we set $\alpha=3$, the models with bilinear projections have very similar or even better performance than the original models with full projections. However, the original MobileNet v2 obtains the best performance compared to the bilinear projection based models even when $\alpha=3$. This is mainly because the extensive use of depth-wise separable convolution in MobileNet v2 is a highly structured module, therefore further using structured bilinear projection for $1\times1$ convolution can hardly increase the representation capacity. The accuracies obtained by both channel pruning (CP) and low-rank expansion (LR) suffer from around 1\% degradation on both datasets. Considering the model sizes illustrated in Table \ref{TB:NTP}, it is quite worth applying our method to deploy CNNs on portable devices with limited memory.

\begin{figure}[h]
\centering
\includegraphics[width=0.4\textwidth]{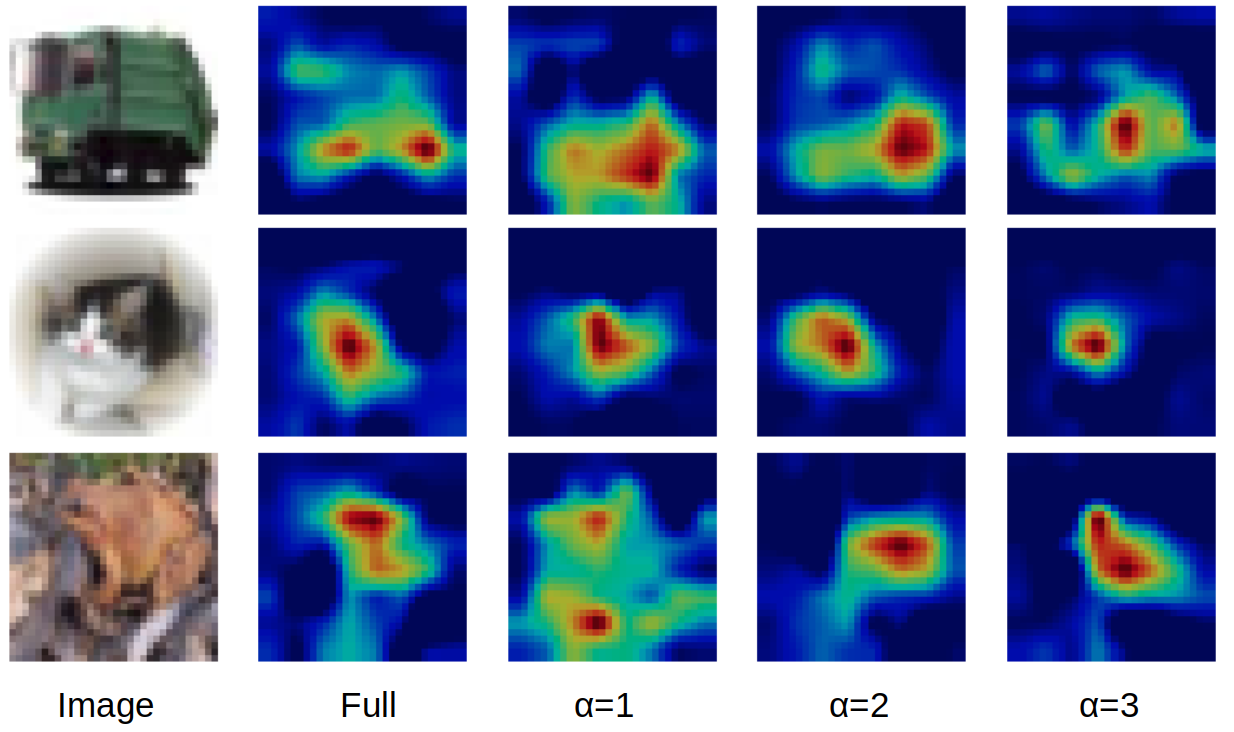}
\caption{The heat maps of ResNet-56 on CIFAR-10 dataset.}
\label{FIG:HEATMAP}
\end{figure}

In Fig. \ref{FIG:HEATMAP}, we illustrate the activation maps from ResNet-56 and its bilinear projection variants using the method proposed in \cite{CVPR16:CAM}. From the heat maps, we can see that when $\alpha=1$ the model obtains comparably scattered activations. However, the models with bilinear projections can almost localize the discriminative regions in images for classification.

\begin{table}[h]
\centering \small
\caption{Classification accuracies on CIFAR-10 test set.}
\label{TB:ACC_CIFAR}
\setlength{\tabcolsep}{4pt}
\renewcommand{\arraystretch}{0.5}
\begin{tabular}{|c|c|c|c|c|c|c|}
\hline
\multirow{2}{*}{Models}   & \multirow{2}{*}{Full}     & \multicolumn{3}{c|}{BP} &\multirow{2}{*}{CP} &\multirow{2}{*}{LR}   \\
\cline{3-5} & & $\alpha=1$ & $\alpha=2$ & $\alpha=3$ &  &\\
\hline
S-VGG      	  & 82.9  & 68.5  & 74.3 & {\bf 83.1}  & 81.7 & 81.8  \\
ResNet-56	  & 92.7  & 83.4  & 89.9 & {\bf 93.0}  & 91.3 & 90.5     \\
SqueezeNet	  & 81.6  & 66.3  & 72.5 & {\bf 81.9}  & - & -      \\
MobileNet v2   & {\bf 80.3} & 70.4 & 74.0  & 78.7   & -  & -       \\
\hline
\end{tabular}
\end{table}

\begin{table}[h]
\centering \small
\caption{Classification accuracies on SVHN test set.}
\label{TB:ACC_SVHN}
\setlength{\tabcolsep}{4pt}
\renewcommand{\arraystretch}{0.5}
\begin{tabular}{|c|c|c|c|c|c|c|}
\hline
\multirow{2}{*}{Models}   & \multirow{2}{*}{Full}     & \multicolumn{3}{c|}{BP}  &\multirow{2}{*}{CP} &\multirow{2}{*}{LR}   \\
\cline{3-5} & & $\alpha=1$ & $\alpha=2$ & $\alpha=3$ & &\\
\hline
S-VGG     & 93.6  & 87.3 & 89.1 & {\bf 93.7}  & 92.3 & 92.2      \\
ResNet-56	  & 97.3  & 93.7  & 95.4 & {\bf 97.3}   & 96.4 & 96.1    \\
SqueezeNet	  & {\bf 92.1}  & 87.4 & 90.5 & 92.0   & - & -      \\
MobileNet v2  & {\bf 94.6}  & 88.6 & 92.3 & 94.4   & - & -    \\
\hline
\end{tabular}
\end{table}

\subsection{Experimental results on image captioning}

\begin{figure*}[h]
\centering
    \begin{minipage}{0.4\textwidth}
        \includegraphics[width=1\textwidth]{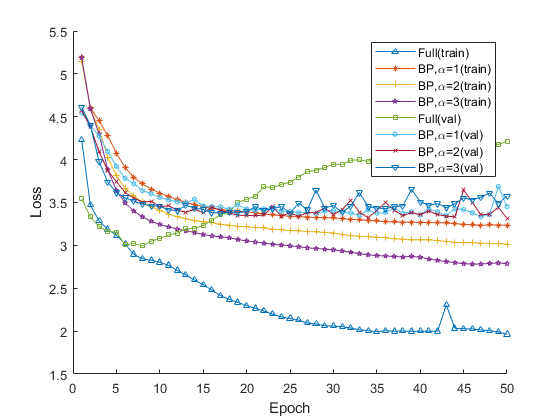}
        \subcaption{Loss.}
    \end{minipage}
    ~ 
    \begin{minipage}{0.4\textwidth}
        \includegraphics[width=1\textwidth]{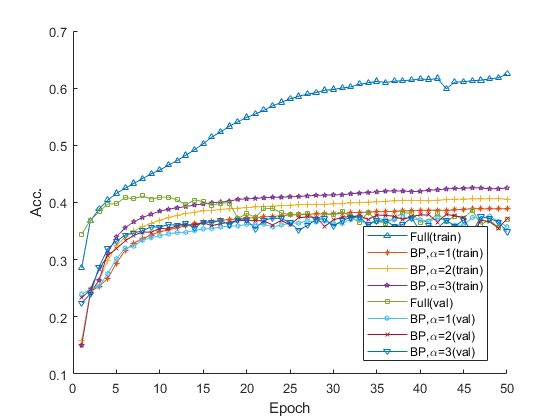}
        \subcaption{Accuracy.}
    \end{minipage}
    \caption{Loss and accuracy curves on Flickr8k dataset.}
\label{FIG:CG}
\end{figure*}

We plotted the loss and accuracy curves of the recurrent image captioning model on Flickr8k in Fig. \ref{FIG:CG}. In the optimisation, the cross-entropy value and accuracy on the training set converge within a few epochs, but these models become over-fitting after that. Applying bilinear projections to reconstruct LSTM and fully-connected layers, there are very minor differences when $\alpha$ changes. This phenomenon signifies even using the smallest size of LSTM, the recurrent model can still have the capability to remember the temporal information over a longer period. A comparison of accuracy and BLEU obtained by bilinear projections and full projections on the test set of Flickr8k is shown in Table \ref{TB:BLEU}. When $\alpha=1$, our efficient method obtains an accuracy of 35.4\% and BLEU of 54.3\%, compared to an accuracy of 40.4\% and BLEU of 57.8\% obtained by the original model. At the same time, using bilinear projections in the embedding layer, two LSTM layers and one dense layer is 1.9x more parameter-efficient than the conventional full projections. Note that the evaluation metric BLEU is not directly related to the sequential classification accuracy. This is because the accuracy per word does not give the score for the entire generated sentence, so for a partially generated sequence, it is non-trivial to balance how good as it is now and the next score as the entire sequence. In fact, when increasing the scaling parameter $\alpha$ from 1 to 3, the model does not have a significant improvement to evaluate the language description.

\begin{table}[h]
\centering \small
\caption{The performance of image caption generation on Flickr8k test set.}
\label{TB:BLEU}
\begin{tabular}{|c|c|c|c|c|c|}
\hline
\multirow{2}{*}{Metrics}   & \multirow{2}{*}{Full}     & \multicolumn{3}{c|}{BP}    \\
\cline{3-5} & & $\alpha=1$ & $\alpha=2$ & $\alpha=3$\\
\hline
Accuracy     & {\bf 40.4}  & 35.4 & 36.7 &  36.8     \\
\hline
BLEU	  & {\bf 57.8}  & 54.3  & 54.9   & 55.4    \\
\hline
\end{tabular}
\end{table}

\subsection{Discussion}
\label{SUBSEC:DISCUSS}

The goal of our proposed method in this paper is to reduce the layer size with a tolerable decrease of accuracy. In conventional architectures, many types of neural layers that contain trainable parameters are basically trained in a back-propagation manner, and nearly all of them can be considered as extensions of dense layers. Our proposed method can reduce the number of trainable parameters to make the whole model parameter-efficient. With a proper scaling up of the intermediate output dimensions, bilinear projections still have sufficient freedom degree to fit complex data distributions.

However, the downside of applying structured mapping to achieve a similar accuracy level is it requires more FLOPs (floating point operations per second) and more GPU memory in the training process. In the design of the layers we use the scaling parameter $\alpha$ to control the output dimensions. When $\alpha=1$ the models with bilinear projections have the same FLOPs and GPU memory with the models of full projection. Scaling up the feature map size by increasing $\alpha$ can improve the model accuracy in general, but incurs more FLOPs and memory usage in the optimisation process. In Table \ref{TB:FLOPS} and Table \ref{TB:MEMORY} we summarised the FLOPs and GPU memory needed for the four network structures, S-VGG, ResNet-56, SqueezeNet and MobileNet v2. From the two tables we can see that when a network becomes very deep (ResNet-56 actually has 190 layers in total), the optimisation is computationally expensive when all convolution layers are implemented with bilinear projections. This constraint indicates that it is necessary to balance the model accuracy and computational resources when optimising complex DNNs. However, it does not affect the deployment of the pre-trained models in most applications, because the system does not need to process very large data batches. Furthermore, the intermediate feature maps, which are used to compute the gradients and update the weights, are unnecessary to consume the GPU memory in the inference procedure.

\begin{table}[h]
\centering \small
\caption{The FLOPs comparisons of the experimented models when setting different values of the scaling parameter $\alpha$.}
\label{TB:FLOPS}
\begin{tabular}{|c|c|c|c|}
\hline
Model   & $\alpha=1$     & $\alpha=2$ & $\alpha=3$   \\
\hline
S-VGG    	& 34K  & 91K & 159K   \\
\hline
ResNet-56	& 153K  & 612K  & 1.4M     \\
\hline
SqueezeNet	& 313K	& 1.2M	& 2.7M	\\
\hline
MobileNet v2 & 353K	 & 4.7M	& 23.0M	 \\
\hline
\end{tabular}
\end{table}

\begin{table}[h]
\centering \small
\caption{The GPU memory comparisons of the experimented models when setting different values of the scaling parameter $\alpha$. The batch size is fixed to 32 and the input image size is $32\times32\times3$ for 32bit float data type.}
\label{TB:MEMORY}
\begin{tabular}{|c|c|c|c|}
\hline
Model   & $\alpha=1$     & $\alpha=2$ & $\alpha=3$   \\
\hline
S-VGG    	& 0.02G  & 0.06G  & 0.14G  \\
\hline
ResNet-56	& 0.52G    &2.09G   & 4.71G     \\ 
\hline
SqueezeNet	& 0.03G	& 0.01G	& 0.32G	\\
\hline
MobileNet v2 & 0.06G	 &0.96G 	& 0.48G	 \\
\hline
\end{tabular}
\end{table}

In real-world applications, we may need complex DNNs that contain hundreds of millions of parameters to learn better representations from the increasing amount of data. In addition to the model compression techniques such as pruning that linearly reduce the model size, we proposed an alternative method for high-efficiency of parameters.  We have conducted the image classification and image captioning experiments, showing that although the DNNs with bilinear projections have much smaller model sizes, the performance is not deteriorated and the accuracy can be even boosted.

\section{Conclusion} \label{SEC:CONCLUSION}

In this paper, we proposed to use bilinear projections to replace traditional unstructured full projections to optimise DNNs. Specifically, we illustrated how to reconstruct fully-connected layer, word embedding layer, convolution layer, and recurrent layer with such projections. This method significantly reduces the number of parameters from $\mathcal{O}(D^2)$ to $\mathcal{O}(2D)$, achieving a sub-linear layer size of deep models. To alleviate the under-fitting problem caused by the lower freedom degree of the structured projection, we properly scaled up the mapping size to keep or even boost the model accuracy. We tested popular CNNs and a recurrent image captioner using bilinear projections on four public datasets, and proved that our proposed microstructure are highly parameter-efficient. 

\section*{Acknowledgement}

We thank the NVIDIA corporation for their kind donation of a Titan Xp GPU card for our experiments.

\bibliographystyle{IEEEtran}
\bibliography{reference}

\begin{thebibliography}{10}
\providecommand{\url}[1]{#1}
\csname url@samestyle\endcsname
\providecommand{\newblock}{\relax}
\providecommand{\bibinfo}[2]{#2}
\providecommand{\BIBentrySTDinterwordspacing}{\spaceskip=0pt\relax}
\providecommand{\BIBentryALTinterwordstretchfactor}{4}
\providecommand{\BIBentryALTinterwordspacing}{\spaceskip=\fontdimen2\font plus
\BIBentryALTinterwordstretchfactor\fontdimen3\font minus
  \fontdimen4\font\relax}
\providecommand{\BIBforeignlanguage}[2]{{%
\expandafter\ifx\csname l@#1\endcsname\relax
\typeout{** WARNING: IEEEtran.bst: No hyphenation pattern has been}%
\typeout{** loaded for the language `#1'. Using the pattern for}%
\typeout{** the default language instead.}%
\else
\language=\csname l@#1\endcsname
\fi
#2}}
\providecommand{\BIBdecl}{\relax}
\BIBdecl

\bibitem{NIPS15:RCNN}
S.~Ren, K.~He, R.~Girshick, and J.~Sun, ``Faster {R-CNN}: Towards real-time
  object detection with region proposal networks,'' in \emph{NIPS}, 2015.

\bibitem{CVPR18:DFN_SEGMENTATION}
C.~Yu, J.~Wang, C.~Peng, C.~Gao, G.~Yu, and N.~Sang, ``Learning a
  discriminative feature network for semantic segmentation,'' in \emph{CVPR},
  2018.

\bibitem{CVPR18:DELS3D}
P.~Wang, R.~Yang, B.~Cao, W.~Xu, and Y.~Lin, ``Dels-3d: Deep localization and
  segmentation with a 3d semantic map,'' in \emph{CVPR}, 2018.

\bibitem{CVPR18:LA_TRACKING}
Q.~Wang, Z.~Teng, J.~Xing, J.~Gao, W.~Hu, and S.~Maybank, ``Learning
  attentions: Residual attentional siamese network for high performance online
  visual tracking,'' in \emph{CVPR}, 2018.

\bibitem{ICANN:LENET}
Y.~LeCun, L.~Jackel, L.~Bottou, A.~Brunot, C.~Cortes, J.~Denker, H.~Drucker,
  I.~Guyon, U.~Muller, E.~Sackinger \emph{et~al.}, ``Comparison of learning
  algorithms for handwritten digit recognition,'' in \emph{Int'l Conf. on
  artificial neural networks}, vol.~60, 1995, pp. 53--60.

\bibitem{CVPR16:INCEPTION}
C.~Szegedy, V.~Vanhoucke, S.~Ioffe, J.~Shlens, and Z.~Wojna, ``Rethinking the
  inception architecture for computer vision,'' in \emph{CVPR}, 2016, pp.
  2818--2826.

\bibitem{CVPR16:RESNET}
K.~He, X.~Zhang, S.~Ren, and J.~Sun, ``Deep residual learning for image
  recognition,'' in \emph{CVPR}, 2016, pp. 770--778.

\bibitem{AAAI17:INCEPTION_RES}
C.~Szegedy, S.~Ioffe, V.~Vanhoucke, and A.~A. Alemi, ``Inception-v4,
  inception-resnet and the impact of residual connections on learning.'' in
  \emph{AAAI}, 2017, pp. 4278--4284.

\bibitem{CVPR15:CG}
O.~Vinyals, A.~Toshev, S.~Bengio, and D.~Erhan, ``Show and tell: A neural image
  caption generator,'' in \emph{CVPR}, 2015, pp. 3156--3164.

\bibitem{AAAI17:RECONSTRUCTION_TRANSLATION}
Z.~Tu, Y.~Liu, L.~Shang, X.~Liu, and H.~Li, ``Neural machine translation with
  reconstruction,'' in \emph{AAAI}, 2017, pp. 3097--3103.

\bibitem{BIO:DL_NER}
M.~Habibi, L.~Weber, M.~Neves, D.~L. Wiegandt, and U.~Leser, ``Deep learning
  with word embeddings improves biomedical named entity recognition,''
  \emph{Bioinformatics}, vol.~33, no.~14, pp. 37--48, 2017.

\bibitem{ACL17:GSN_QA}
W.~Wang, N.~Yang, F.~Wei, B.~Chang, and M.~Zhou, ``Gated self-matching networks
  for reading comprehension and question answering,'' in \emph{ACL}, 2017, pp.
  189--198.

\bibitem{NC:LSTM}
S.~Hochreiter and J.~Schmidhuber, ``Long short-term memory,'' \emph{Neural
  computation}, vol.~9, no.~8, pp. 1735--1780, 1997.

\bibitem{NIPS15:LWC}
S.~Han, J.~Pool, J.~Tran, and W.~Dally, ``Learning both weights and connections
  for efficient neural network,'' in \emph{NIPS}, 2015, pp. 1135--1143.

\bibitem{NIPS16:LSS}
W.~Wen, C.~Wu, Y.~Wang, Y.~Chen, and H.~Li, ``Learning structured sparsity in
  deep neural networks,'' in \emph{NIPS}, 2016, pp. 2074--2082.

\bibitem{ARXIV:AT}
S.~Zagoruyko and N.~Komodakis, ``Paying more attention to attention: Improving
  the performance of convolutional neural networks via attention transfer,''
  \emph{arXiv preprint arXiv:1612.03928}, 2016.

\bibitem{ARXIV:NT}
H.~Hu, R.~Peng, Y.-W. Tai, and C.-K. Tang, ``Network trimming: A data-driven
  neuron pruning approach towards efficient deep architectures,'' \emph{arXiv
  preprint arXiv:1607.03250}, 2016.

\bibitem{ICCV17:CP}
Y.~He, X.~Zhang, and J.~Sun, ``Channel pruning for accelerating very deep
  neural networks,'' in \emph{ICCV}, 2017, pp. 1389--1397.

\bibitem{ARXIV:CP}
V.~Lebedev, Y.~Ganin, M.~Rakhuba, I.~Oseledets, and V.~Lempitsky, ``Speeding-up
  convolutional neural networks using fine-tuned cp-decomposition,''
  \emph{arXiv preprint arXiv:1412.6553}, 2014.

\bibitem{ARXIV:LR}
M.~Jaderberg, A.~Vedaldi, and A.~Zisserman, ``Speeding up convolutional neural
  networks with low rank expansions,'' \emph{arXiv preprint arXiv:1405.3866},
  2014.

\bibitem{ICCV15:HTC}
W.~Chen, J.~Wilson, S.~Tyree, K.~Weinberger, and Y.~Chen, ``Compressing neural
  networks with the hashing trick,'' in \emph{ICCV}, 2015, pp. 2285--2294.

\bibitem{CVPR16:QCNN}
J.~Wu, C.~Leng, Y.~Wang, Q.~Hu, and J.~Cheng, ``Quantized convolutional neural
  networks for mobile devices,'' in \emph{CVPR}, 2016, pp. 4820--4828.

\bibitem{ARIXV:CVQ}
Y.~Gong, L.~Liu, M.~Yang, and L.~Bourdev, ``Compressing deep convolutional
  networks using vector quantization,'' \emph{arXiv preprint arXiv:1412.6115},
  2014.

\bibitem{ARXIV:SQUEEZENET}
F.~N. Iandola, S.~Han, M.~W. Moskewicz, K.~Ashraf, W.~J. Dally, and K.~Keutzer,
  ``Squeezenet: Alexnet-level accuracy with 50x fewer parameters and< 0.5 mb
  model size,'' \emph{ICLR}, 2016.

\bibitem{ARIXV:MOBILENET}
A.~G. Howard, M.~Zhu, B.~Chen, D.~Kalenichenko, W.~Wang, T.~Weyand,
  M.~Andreetto, and H.~Adam, ``Mobilenets: Efficient convolutional neural
  networks for mobile vision applications,'' \emph{arXiv preprint
  arXiv:1704.04861}, 2017.

\bibitem{CVPR18:MOBILENET}
M.~Sandler, A.~Howard, M.~Zhu, A.~Zhmoginov, and L.-C. Chen, ``Mobilenetv2:
  Inverted residuals and linear bottlenecks,'' in \emph{CVPR}, 2018, pp.
  4510--4520.

\bibitem{CVPR17:DENSENET}
G.~Huang, Z.~Liu, K.~Q. Weinberger, and L.~van~der Maaten, ``Densely connected
  convolutional networks,'' in \emph{CVPR}, 2017, pp. 4700--4708.

\bibitem{CVPR18:CONDENSENET}
G.~Huang, S.~Liu, L.~van~der Maaten, and K.~Q. Weinberger, ``Condensenet: An
  efficient densenet using learned group convolutions,'' pp. 2752--2761, 2018.

\bibitem{ICCV15:CIR}
Y.~Cheng, F.~X. Yu, R.~S. Feris, S.~Kumar, A.~Choudhary, and S.-F. Chang, ``An
  exploration of parameter redundancy in deep networks with circulant
  projections,'' in \emph{ICCV}, 2015, pp. 2857--2865.

\bibitem{JBO:HYPER_IMAGE}
G.~Lu and B.~Fei, ``Medical hyperspectral imaging: a review,'' \emph{Journal of
  biomedical optics}, vol.~19, no.~1, pp. 1--9, 2014.

\bibitem{IJCV:FV}
J.~S{\'a}nchez, F.~Perronnin, T.~Mensink, and J.~Verbeek, ``Image
  classification with the fisher vector: Theory and practice,'' \emph{Int'l
  journal of computer vision}, vol. 105, no.~3, pp. 222--245, 2013.

\bibitem{CVPR10:VLAD}
H.~J{\'e}gou, M.~Douze, C.~Schmid, and P.~P{\'e}rez, ``Aggregating local
  descriptors into a compact image representation,'' in \emph{CVPR}, 2010, pp.
  3304--3311.

\bibitem{SOC06:FJLT}
N.~Ailon and B.~Chazelle, ``Approximate nearest neighbors and the fast
  johnson-lindenstrauss transform,'' in \emph{ACM Symposium on Theory of
  computing}, 2006, pp. 557--563.

\bibitem{KDD11:FLSH}
A.~Dasgupta, R.~Kumar, and T.~Sarl{\'o}s, ``Fast locality-sensitive hashing,''
  in \emph{ACM SIGKDD}, 2011, pp. 1073--1081.

\bibitem{ICML14:CBE}
F.~Yu, S.~Kumar, Y.~Gong, and S.-F. Chang, ``Circulant binary embedding,'' in
  \emph{ICML}, 2014, pp. 946--954.

\bibitem{CVPR13:CTE}
J.~Revaud, M.~Douze, C.~Schmid, and H.~J{\'e}gou, ``Event retrieval in large
  video collections with circulant temporal encoding,'' in \emph{CVPR}, 2013,
  pp. 2459--2466.

\bibitem{STOC97:TA}
J.~M. Kleinberg, ``Two algorithms for nearest-neighbor search in high
  dimensions,'' in \emph{STOC}, vol.~97, 1997, pp. 599--608.

\bibitem{PR:MRSTM}
C.~Hou, F.~Nie, C.~Zhang, D.~Yi, and Y.~Wu, ``Multiple rank multi-linear svm
  for matrix data classification,'' \emph{Pattern Recognition}, vol.~47, no.~1,
  pp. 454--469, 2014.

\bibitem{TIP:MRR}
C.~Hou, F.~Nie, D.~Yi, and Y.~Wu, ``Efficient image classification via multiple
  rank regression,'' \emph{IEEE Trans. on Image Processing}, vol.~22, no.~1,
  pp. 340--352, 2012.

\bibitem{ARXIV:VGG}
K.~Simonyan and A.~Zisserman, ``Very deep convolutional networks for
  large-scale image recognition,'' \emph{arXiv preprint}, 2014.

\bibitem{ECCV16:RESNET}
K.~He, X.~Zhang, S.~Ren, and J.~Sun, ``Identity mappings in deep residual
  networks,'' in \emph{ECCV}, 2016, pp. 630--645.

\bibitem{CVPR15:BICNN}
T.-Y. Lin, A.~RoyChowdhury, and S.~Maji, ``Bilinear cnn models for fine-grained
  visual recognition,'' in \emph{CVPR}, 2015, pp. 1449--1457.

\bibitem{NIPS13:WORD2VEC}
T.~Mikolov, I.~Sutskever, K.~Chen, G.~S. Corrado, and J.~Dean, ``Distributed
  representations of words and phrases and their compositionality,'' in
  \emph{NIPS}, 2013, pp. 3111--3119.

\bibitem{ARXIV:LSTM}
W.~Zaremba and I.~Sutskever, ``Learning to execute,'' \emph{arXiv preprint
  arXiv:1410.4615}, 2014.

\bibitem{HLT10:FLICKR8K}
C.~Rashtchian, P.~Young, M.~Hodosh, and J.~Hockenmaier, ``Collecting image
  annotations using amazon's mechanical turk,'' in \emph{NAACL HLT Workshop},
  2010, pp. 139--147.

\bibitem{ACL02:BLEU}
K.~Papineni, S.~Roukos, T.~Ward, and W.-J. Zhu, ``Bleu: a method for automatic
  evaluation of machine translation,'' in \emph{ACL}, 2002, pp. 311--318.

\bibitem{CVPR16:CAM}
B.~Zhou, A.~Khosla, A.~Lapedriza, A.~Oliva, and A.~Torralba, ``Learning deep
  features for discriminative localization,'' in \emph{CVPR}, 2016, pp.
  2921--2929.

\end{thebibliography}

\end{document}